\documentclass[runningheads]{llncs}
\usepackage{graphicx}
\usepackage{tikz}
\newcommand*\circled[1]{\tikz[baseline=(char.base)]{
            \node[shape=circle,draw,inner sep=1pt] (char) {#1};}}
\newcommand{\fakeparagraph}[1]{\smallskip\noindent\textbf{#1.}}
\begin{document}
\title{Explainable AI meets Healthcare: 
A Study on Heart Disease Dataset}
\author{Devam Dave\inst{1} \and Het Naik\inst{1} \and Smiti Singhal\inst{1} \and Pankesh Patel\inst{2}}
\institute{
Pandit Deendayal Petroleum University, Gandhinagar, India. \\
\email{\{devam.dce18, het.nce18, smiti.sce18\}@sot.pdpu.ac.in} 
\and 
Confirm SFI Centre for Smart Manufacturing, Data Science Institute, \\ National University of Ireland Galway \\
\email{pankesh.patel@insight-centre.org} 
}
\maketitle
\begin{abstract}
With the increasing availability of structured and unstructured data and the swift progress of analytical techniques, Artificial Intelligence (AI) is bringing a revolution to the healthcare industry. With the increasingly indispensable role of AI in healthcare, there are growing concerns over the lack of transparency and explainability in addition to potential bias encountered by predictions of the model. This is where Explainable Artificial Intelligence (XAI) comes into the picture. XAI increases the trust placed in an AI system by medical practitioners as well as AI researchers, and thus, eventually, leads to an increasingly widespread deployment of AI in healthcare. 

In this paper, we present different interpretability techniques. The aim is to enlighten practitioners on the understandability and interpretability of explainable AI systems using a variety of techniques available which can be very advantageous in the health-care domain. Medical diagnosis model is responsible for human life and we need to be confident enough to treat a patient as instructed by a black-box model. Our paper contains examples based on the heart disease dataset and elucidates on how the explainability techniques should be preferred to create trustworthiness while using AI systems in healthcare.

\end{abstract}
\section{Introduction}
For healthcare applications where explanation of the inherent logic is important for people who make decisions, machine learning’s lack of explainability restricts the wide-scale deployment of AI.  If AI cannot explain itself in the domain of healthcare, then its risk of making a wrong decision may override its advantages of accuracy, speed and decision-making efficacy. This would, in turn, severely limit its scope and utility. Therefore, it is very important to look at these issues closely. Standard tools must be built before a model is deployed in the healthcare domain. One such tool is explainability (or Explainable AI). The rationale behind the use of Explainable AI techniques is to increase transparency, result tracing and model improvement~\cite{healthintro}. For instance, they explain why someone is categorized as ill or otherwise. This would increase the trust level of medical practitioners to rely on AI. Eventually, XAI can be integrated into smart healthcare systems involving IoT, Cloud computing and AI primarily used in the areas of cardiology, cancer and neurology~\cite{major}. These smart healthcare systems can then be used for diagnosing diseases and selection of the appropriate treatment plan. In this paper, we look at some examples of various XAI techniques carried out on the Heart Disease Dataset from UCI \footnote{Heart Disease Dataset: ~\url{https://www.kaggle.com/cherngs/heart-disease-cleveland-uci}} along with the use cases related to the technique.~\cite{xaibook}

\fakeparagraph{Objectives} The objective of this paper is to study and utilize different explainable AI techniques in the healthcare sector, as it gives transparency, consistency, fairness and trust to the system. The particulars for the objectives are : 
\begin{itemize}
    \item To study feature-based and example-based explainable AI techniques using the heart disease dataset.
    \item To draw out the inferences from the results of these techniques and conclude  selection of one technique over the other for a particular area of healthcare.
\end{itemize}

We have worked on various techniques that give explanation of outcomes given by the black box models. The paper gives insights on how these techniques are advantageous in different conditions. Moreover, the different approaches followed by them are studied.

\fakeparagraph{Outline} The following is the structure of the paper. Section 2 gives an overview
of our approach, consisting different phases of Machine Learning~(such as Model training,  Model deployment) and explainable AI.  A brief description of the dataset along with the explanation of features is presented in Section 3. Section 4 talks about the feature based techniques LIME and SHAP giving explanation of the methods in detail with the support of examples. We describe various example based techniques in the next Section 5. It gives insights on the different techniques available in the library alibi and demonstrates their importance in Explainable AI.  Lastly, Section 6 discusses the findings of the entire paper and concludes the work.

\section{Our approach} 

\begin{figure}[!ht]
  \centering
  \includegraphics[scale=0.4]{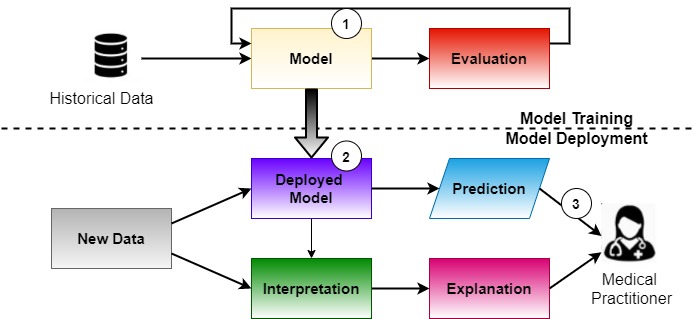}
  \caption{A ML Life-Cycle in Conjunction with XAI}
  \label{fig:our-approach}
\end{figure}

The objective of our research is to present our early research work on building Explainable AI-based approach for healthcare applications. Figure~\ref{fig:our-approach} presents an ML life-cycle~\cite{xaibook} in conjunction with XAI methods to obtain greater benefit from AI-based black box models deployed in the healthcare domain.

\fakeparagraph{Model training} ML algorithms use the historical healthcare data collected for the purposes of training where they attempt to learn latent patterns and relationships from data without hardwiring the fixed rules~(Circled~\circled{1} in Figure~\ref{fig:our-approach}). We use the Heart Disease Dataset from the UCI ML Repository. This dataset contains 70+ features. Its objective is to detect the existence of heart disease in patients.  The training leads to the generation of models, which are deployed in the production environments~(Circled~\circled{2} in Figure~\ref{fig:our-approach}).  We use the XGBoost~(eXtreme Gradient Boosting) algorithm for training, which is an implementation of the gradient boosted decision trees developed for performance as well as speed.

XGBoost is a decision-tree based machine learning algorithm based on a boosting framework. Outliers don’t have a significant impact on the performance of XGBoost. There is no requirement for carrying out feature engineering in XGBoost either. This is one of the only algorithms that can derive feature importance, which is an integral part of Explainable AI. XGBoost is highly suitable for any kind of classification problem and therefore, appropriate for our paper and project.

\fakeparagraph{Model deployment and interpretability} The objective of implementing explainable AI techniques along with the deployed model is to interpret how the prediction results are derived. The data is supplied to the trained model and explainable AI module. By implementing explainable AI techniques, it allows us to provide explanations along with prediction results~(Circled~\circled{3} in Figure~\ref{fig:our-approach}). The explanation can be consumed by medical practitioners to validate the predictions made by the AI models. Clinical records along with explanation can be used to generate deeper insights and recommendations. Section~\ref{sec:feature-based-tech} and Section~\ref{sec:example-based-tech} present the explanation generated by various explainable AI techniques.

Before we present various explainable AI techniques and their results, the next section~(Section~\ref{sec:dataset}) presents our heart disease case study and describe the dataset.

\section{Dataset Description}\label{sec:dataset}

Heart disease is among the biggest causes of deaths in the entire world, its mortality rate will even increase in the Post-Covid era as many heart problems arise due to it. Prediction of a heart disease is one of  the most important area in the healthcare sector. Heart disease refers to blockage or narrowing down of blood vessels, which can lead to chest pain or heart attack. 
The Heart Disease Cleveland UC Irvine dataset is based on prediction if a person has heart disease or not based on 13 attributes\footnote{Heart disease dataset : \url{https://www.kaggle.com/cherngs/heart-disease-cleveland-uci}}
. It is re-processed from the original dataset having 76 features. In the following, we describe 13 features briefly:

\begin{enumerate}
    \item \textit{age}: An integer value signifying the age of an individual. 
    \item \textit{sex}: \textbf{0} for female, \textbf{1} for male. 
    \item \textit{cp}: \textit{cp} stands for the Chest Pain Type and ranges from \textbf{0-3} depending upon the symptoms experienced by a patient. 
    They are classified using the symptoms experienced by a patient. The three main symptoms of angina are:
    \begin{itemize}
        \item Substernal chest discomfort
        \item Provoked by physical exertion or emotional stress
        \item Rest and/or nitroglycerine relieves 
    \end{itemize}
According to the symptoms experienced, chest pain types are classified in the following ways:
    \begin{enumerate} 
        \item \textbf{0} - Typical Angina: Experiencing all three symptoms
        \item \textbf{1} - Atypical Angina: Experiencing any two symptoms
        \item \textbf{2} - Non-Anginal Pain: Experiencing any one symptom
        \item \textbf{3} - Asymptomatic Pain: Experiencing none of the symptoms mentioned above
    \end{enumerate}
If \textit{cp} is high, less exercise should be performed and sugar and cholesterol level of the body should be maintained

    \item \textit{trestbps}: \textit{trestbps} shows the resting blood pressure calculated in units of millimeters in mercury (mmHg). The ideal blood pressure is 90/60mmHg to 120/80mmHg. High blood pressure is considered to be anything above 140/90mmHg. If trestbps is high, less exercise should be performed and if \textit{trestbps} is less , respective medicine should be taken. 

    \item \textit{chol}: \textit{chol} represents the cholesterol levels of an individual. A high cholesterol level can lead to blockage of heart vessels. Ideally, cholesterol levels should be below 170mg/Dl for healthy adults. If \textit{chol} is high, normal exercise should be performed, and less oily food should be eaten.

    \item \textit{fbs}: \textit{fbs} represents Fasting Blood Sugar levels of a patient, by gauging the amount of glucose present in the blood. A blood sugar level below 100 mmol/L is considered to be normal. 
    \begin{enumerate}
        \item \textbf{1} signifies that the patient has a blood sugar level in excess of 120mmol/L
        \item \textbf{0} signifies that the patient has a blood sugar level lower than 120mmol/L
    \end{enumerate}
     and  If \textit{fbs} is high, dash diet should be taken, and frequent intake of food in less amount should be taken 
     
    \item \textit{restecg}: \textit{restecg} depicts the electrocardiograph results of a patient. \textit{restecg} ranges between \textbf{0-2}.
    \begin{enumerate}
        \item \textbf{0} - Normal results in ECG
        \item \textbf{1} - The ECG Results have a ST-T wave abnormality
        \item \textbf{2} - The ECG Results show a probable or definite left ventricular hypertrophy by Estes’ criteria
    \end{enumerate}

    If \textit{restecg} is high, moderate exercise should be performed, dash diet should be taken, and frequent intake of food in less amount should be taken.

    \item \textit{thalach}: \textit{thalach} shows the maximum heart rate of an individual using a Thallium Test. A Thallium test is an unconventional method for checking heart disease. It is carried out by injecting a small amount of radioactive substance (Thallium in this case) into the bloodstream of an individual, while he/she is exercising. Using a special camera, the blood flow and the pumping of the heart can be determined. \textit{thalach} denotes the maximum heart rate achieved during this Thallium test. If \textit{thalach} is low, proper exercise should be performed

    \item \textit{exang}: \textit{exang} is a feature that reveals whether a patient has exercise induced angina (signified by \textbf{1}) or not (signified by \textbf{0}). Exercise induced angina is a kind of angina that is triggered by physical activity and exertion due to an increase in the demand of oxygen. 

    \item \textit{oldpeak}: \textit{oldpeak} is the amount of depression of the ST Wave in the case of a patient having a value of \textbf{1} in \textit{restecg} (ST-T Wave abnormality found in ECG Results). This peak is induced by exercise and is measured relative to the results at rest. If \textit{oldpeak} is high, less exercise should be performed, and dash diet should be taken

    \item \textit{slope}: \textit{slope} is also concerned with the ST Wave in ECG Results. 
    \begin{enumerate}
        \item \textbf{0} signifies an upward slope in the ST Wave
        \item \textbf{1} signifies that the ST Wave is flat
        \item \textbf{2} signifies a downward slope in the ST Wave
    \end{enumerate}
    If the \textit{slope} is high, respective medicines should be taken, and dash diet should be taken.

    \item \textit{ca}: The heart has 3 main vessels responsible for blood flow. An angiography is carried out and because of the dye, the unblocked vessels show up on a X-Ray. Ideally, three vessels should be visible on the X-Ray as this would mean that none of the vessels are blocked. If \textit{ca} is high, angioplasty should be performed for the treatment of blocked vessels, at the later stage stent should be put if required, and bypass surgery should be performed in the worst case scenari\textbf{0}
    \item \textit{thal}: \textit{thal} denotes the results of Thallium test of an individual. 
    \begin{enumerate}
        \item \textbf{0} denotes that the results were normal.
        \item \textbf{1} denotes a fixed defect.
        \item \textbf{2} denotes a reversible defect.
    \end{enumerate}
    Here, defect symbolises an obstruction in optimum blood flow. Thallium test is done in a physically exerted state. Fixed defect conveys a defect that stays even after the body is at rest. On the other hand, reversible defect is a defect that passes away as the body relaxes. 
\end{enumerate}

\begin{figure}[!ht]
  \centering
  \includegraphics[scale=0.4]{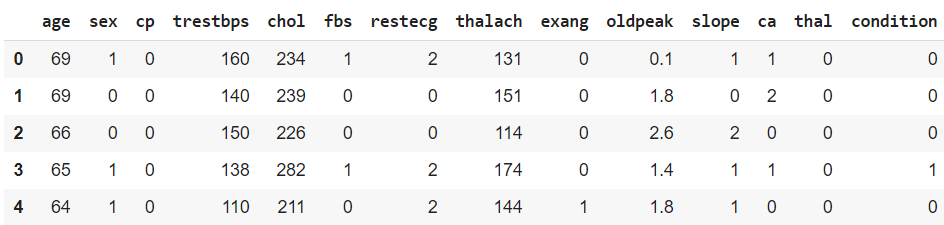}
  \caption{Example instances of dataset. }
  \label{fig:heart disease dataset}
\end{figure}

Figure~\ref{fig:heart disease dataset} depicts example instances of the heart disease dataset. Using this dataset, we present the explanation generated by various explainable AI techniques in Section~\ref{sec:feature-based-tech} and Section~\ref{sec:example-based-tech}.

\section{Feature-Based Techniques}\label{sec:feature-based-tech}

This section presents Feature-based model explainability techniques, which denote how
much the input features contribute to a model’s output. There are many Feature-based methods
available such as permutation Feature Importance, Partial Dependence Plots (PDPs), Individual Conditional Expectation (ICE) plots, Accumulated Local Effects (ALE) Plot, Global
surrogate models, Local Interpretable Model-agnostic Explanations (LIME) and Shapley
Additive Explanations (SHAP). We discuss these methods in this section.

\subsection{Local Interpretable Model-Agnostic Explanations (LIME)}\label{sec:lime}
LIME ~\cite{anchors:aaai18} is a technique that does not try to explain the whole model, instead  LIME tries to understand the model by perturbing the input of data samples and comprehending how the predictions change.  LIME enables local model interpretability. A single data sample is modified adjusting some feature values and the resultant output impact is observed. This is often linked to what human interests are when the output of a model is observed. 

Let us see an example on heart disease dataset to understand the concept further.
\begin{verbatim}
print('Reference:', y_test.iloc[target])
print('Predicted:', predictions[target])
exp.explain_instance(X_test.iloc[target].values,
                     xgb_array.predict_proba)
\end{verbatim}

\begin{figure}[!ht]
  \centering
  \includegraphics[scale=0.2]{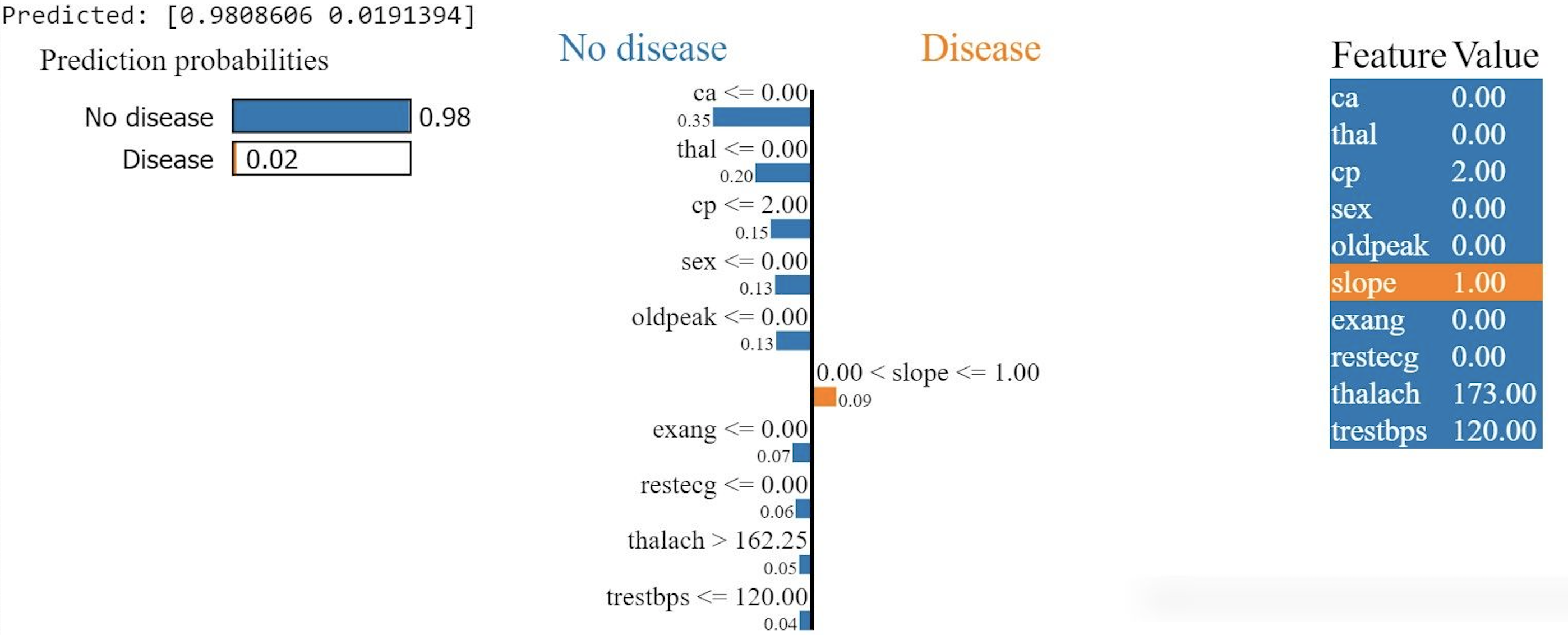}
  \caption{Explanations generated by LIME}
  \label{fig:lime}
\end{figure}

The left part of Figure~\ref{fig:lime} shows the prediction probabilities made by the XGBoost model for the classes ‘\texttt{No disease}’~(\texttt{98\%}) and \texttt{‘Disease’} (\texttt{2\%}).  The middle part  gives the features along with their weights which impact the prediction probabilities. For instance,  we notice if the \texttt{cp} value, in the range of $0.00$ < \texttt{cp} <= $2.00$, contributes towards not having the disease. The right part indicates the actual value of a particular feature for a specific local instance.

\subsection{SHapley Additive exPlanations (SHAP)}\label{sec:shap}
{SHAP~\cite{lundberg2017unified}} method increases the transparency of the model. It can be understood well using the game theory concept from where the concept arises. The intuition is as follows. A prediction can be explained by presuming that each feature value of the instance is a “player” in a game where the pay-out is the prediction itself. Stemming from a method of coalitional game theory, Shapley values tell us how to distribute the ``pay-out'' fairly among all the features. 

Coined by Shapley in 1953, the Shapley value is a method to assign pay-outs to players based on their individual contribution to the overall pay-out. Players collaborate in an alliance and reap some profits from this collaboration. With its origins in the coalitional game theory, Shapley values point us in the direction of fair distribution of ``pay-out'' amongst all players. This theory can be employed in ML predictions and interpretability. A model prediction is explained by equating each feature value of a dataset instance to a ``player'' in a game wherein the prediction itself is the “pay-out.” The Shapley values throw light on how to distribute the “pay-out” fairly between all features. The ``game'' is the task of prediction for a lone dataset instance. The ``gain'' is the difference between the actual prediction for the single instance and the average prediction for all instances. Finally, the “players” are the instance feature values which cooperate with each other to split the gains (equals to predicting a certain value). 

SHAP sets a mean prediction(base value) of the model and identifies the relative contribution of every feature to the deviation of the target from the base. It can give both local as well as global explanations. Let us consider the local explanations first. 

\fakeparagraph{Local explanation} We will execute on several instances in order to show how the SHAP values behave as a local explanation method. Below given are the example of 3 particular instances which are randomly selected on which the method is to be explained.  Each case results into its own set of SHAP values from which we get to know why a case received the particular prediction and the contributions of the predictors.
\begin{verbatim}
shap.force_plot(explainer.expected_value,
                shap_values[instance_1,:], 
                X_test.iloc[instance_1,:])
\end{verbatim}

\begin{figure}[!ht]
  \centering
  \includegraphics[scale=0.4]{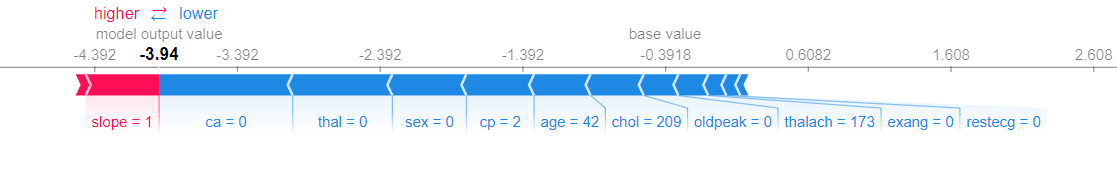}
  \caption{Local Explanation generated by SHAP force plot}
  \label{fig:shap1}
\end{figure}
\begin{verbatim}
shap.force_plot(explainer.expected_value, 
                shap_values[instance_2,:], 
X_test.iloc[instance_2,:])
\end{verbatim}

\begin{figure}[!ht]
  \centering
  \includegraphics[scale=0.4]{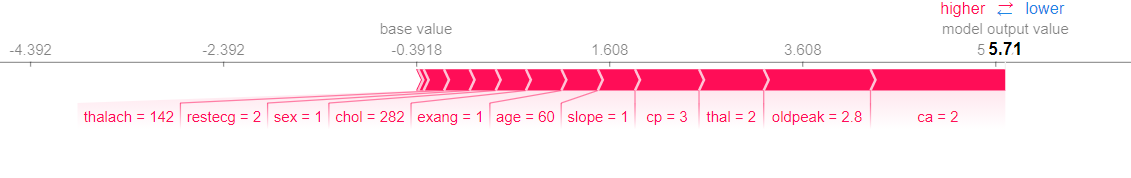}
  \caption{Local Explanation generated by SHAP force plot}
  \label{fig:shap2}
\end{figure}
\begin{verbatim}
shap.force_plot(explainer.expected_value,
                shap_values[instance_3,:], 
X_test.iloc[instance_3,:])
\end{verbatim}

\begin{figure}[!ht]
  \centering
  \includegraphics[scale=0.39]{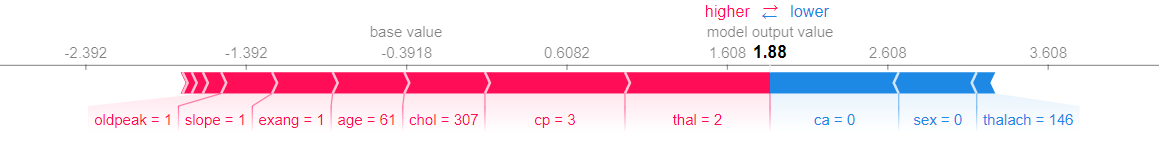}
  \caption{Local Explanation generated by SHAP force plot}
  \label{fig:shap3}
\end{figure}
Using SHAP, we have seen two extreme cases in the same fashion as LIME. Now, we look at a local explanation that isn't as extreme. We see a good mix of Red and Blue feature values from the figure shown above. Looking at both cases individually:

1. The most important feature values influencing the model to make the decision that the patient has a heart disease are \texttt{thal}= $2$ (Showing that there are some defects in the blood supply and the quality of heart cells of the patient), \texttt{cp}= $3$ (Showing that the type of chest pain is asymptomatic. This may seem counterintuitive, but asymptomatic chest pain is actually the most severe out of the four types, which leads the model to believe that the patient has a heart disease.)  and \texttt{chol} = $307$ (High cholestrol leads to blockage of blood vessels of the heart and decreases the overall blood flow in and around the heart). 

2. The most important feature value influencing the model to make the decision that the patient does not have a heart disease are \texttt{ca} = $0$. This is perhaps the most important feature, and from this we realise that none of the vessels of the patient are blocked.  

\fakeparagraph{Global explanation} The collective SHAP values got shows how much each feature contributes, how it contributes i.e. positively or negatively to the final prediction. There are a number of types of plots which can show global explanation as shown below.
\begin{verbatim}
shap.force_plot(explainer.expected_value, 
                shap_values[:1000,:],
                X_test.iloc[:1000,:])
\end{verbatim}

\begin{figure}[ht]
  \centering
  \includegraphics[scale=0.25]{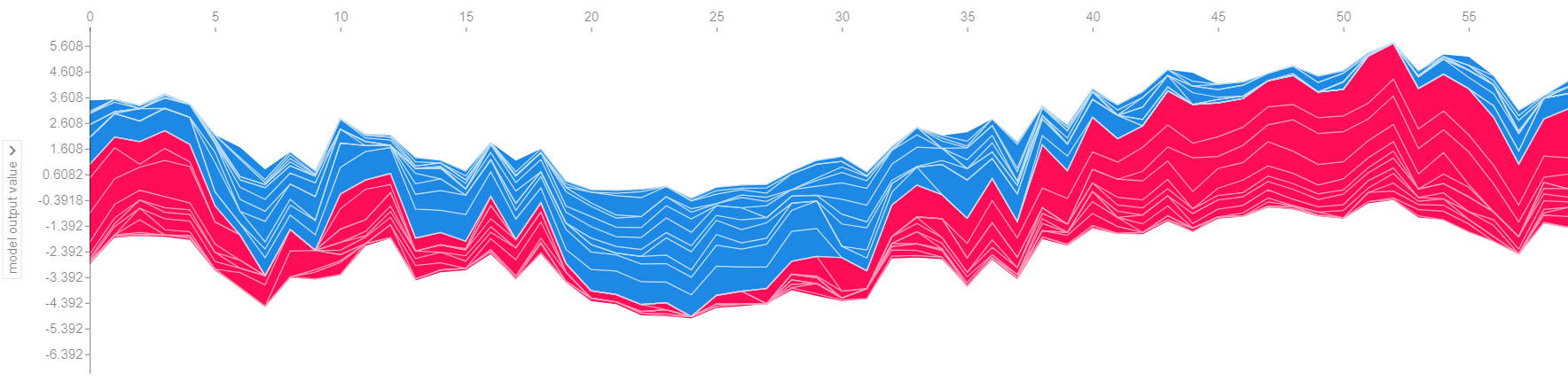}
  \caption{Global Explanations generated by SHAP force plot}
  \label{fig:shap_global}
\end{figure}

Fig \ref{fig:shap_global} is a global explanation of the predictions of the model. -0.3918 is the base value as obtained using the SHAP Values. This means that if the total value is more than -0.3918, it signifies that the patient has the disease and if it is less than -0.3918, it signifies that the patient does not have the disease. The blue part of the graph pushes the prediction lower, and the red part is responsible for increasing it. This means that the instances in which there are a lot more red colored features will usually be $1$ (having a disease) and vice versa.

\begin{verbatim}
 shap.summary_plot(shap_values, X_test)
\end{verbatim}

\begin{figure}[!ht]
  \centering
  \includegraphics[scale=0.7]{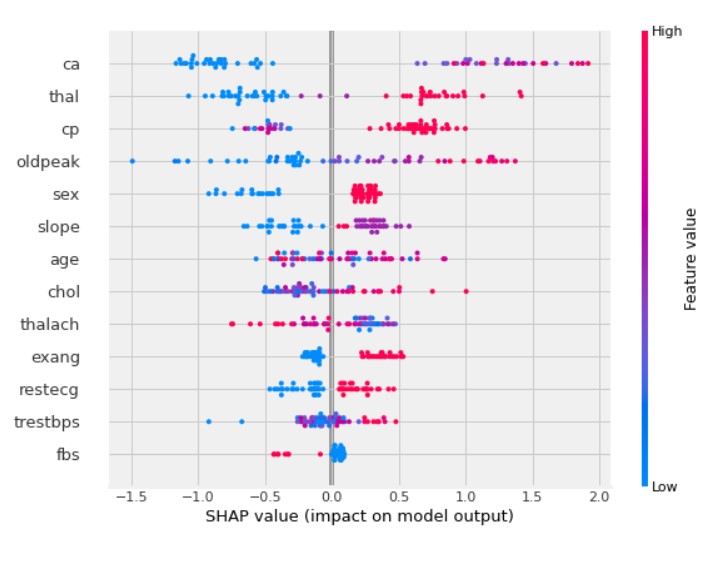}
  \caption{Explanations generated by SHAP summary plot}
  \label{fig:shap_summary}
\end{figure}

By the scatter plot graph shown in Figure \ref{fig:shap_summary}, we have visualised the effects of the features on the prediction at different values. The features at the top contribute more to the model than the bottom ones and thus have high feature importance. The color represents the value of the feature. (Blue meaning low, purple meaning the median value and red meaning high). For example, in \texttt{ca}, we see that when the dots are blue, the shap value is negative and when the dots are red and purple, the shap values are mostly positive. This signifies that when no vessels are blocked, chances of disease are low but as the number of vessels blocked increases, so does the chances of having a disease.

\begin{verbatim}
shap.dependence_plot(ind='thalach', interaction_index='ca',
                     shap_values=shap_values, 
                     features=X_test,  
                     display_features=X_test)
                    
\end{verbatim}

\begin{figure}[!ht]
  \centering
  \includegraphics[scale=0.7]{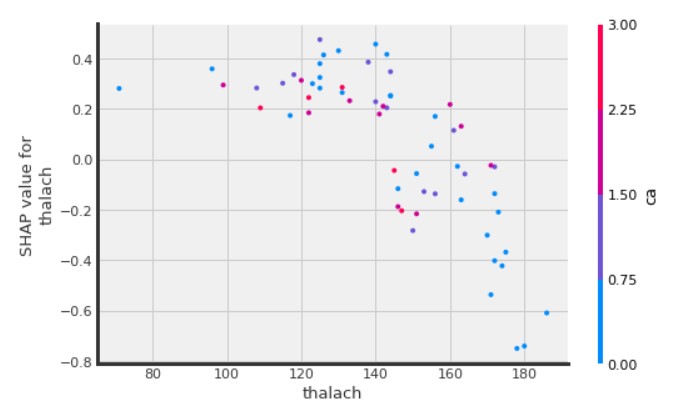}
  \caption{Explanations generated by SHAP scatter plot}
  \label{fig:shap}
\end{figure}

This scatter plot in Figure \ref{fig:shap} shows us the relation between two features, as well as the SHAP Values. The X-axis shows us \texttt{thalach} (maximum heart rate), the Y-axis shows us the SHAP Values and the color of each dot shows us the value of \texttt{ca}. More often than not, the shap values are low when the value of \texttt{ca} is low. There is also a slight trend of the shap values decreasing as the value of \texttt{thalach} increases.

\section{Example-Based Techniques}\label{sec:example-based-tech}

This section presents Example-based model explainability techniques, which denote how
much an instance contributes to a model’s output. There are many Example-based methods\footnote{Alibi example based techniques:~\url{https://github.com/SeldonIO/alibi}}
available such as Anchors, Counterfactuals, Counterfactuals guided by Prototypes, Contrastive Explanation Method (CEM), Kernel Shap, Tree Shap and Integrated Gradients. We discuss these methods in this section.

\subsection{Anchors}\label{sec:anchors}
Anchors are the local-limit sufficient conditions of certain features at which the model gives a high precision prediction \cite{anchors:aaai18}. In anchors, the features are narrowed down to certain conditions (i.e. anchors), which gives the prediction. Anchors is based on the algorithm which works for model-agnostic approach for classification models of text, image and tabular data. Anchor takes into account the global dataset and then gives the anchor feature-values. It is based on \texttt{If-Then} rules for finding the features of the input instance responsible for the prediction which makes it reusable and clear for which other instances it is valid.

Anchors are pretty much similar to LIME as they both provide local explanations linearly, but the problem with LIME is that it only covers the local region which might give overconfidence in the explanation and might not fit for an instance which is not trained. LIME is only able to describe a particular instance which means that, when new real world data is given as input which is unseen, it may give confusing or unfaithful explanations. This is because it only learns from a linear decision boundary that best explains the model. This is where Anchors has an advantage over LIME. Anchors generate a local region that is a better construction of the dataset to be explained. The boundaries are clear and cover the area such that it is best adapted to the model's behaviour. If the same perturbation space is given to both LIME and Anchors, the latter will build a more valid region of instances and which better describes the model’s behaviour. Moreover, their computation time is less as compared to SHAP. Thus, Anchors are a very good choice to validate the model’s behaviour and find its decision rule.

The major drawback of Anchors is the fact that they provide a set of rules which are not only easily interpretable and if they become excessively specific by including a lot of feature predicates, the area of coverage to explain the observations decreases to a great extent.

\begin{figure}[ht]
\centering
\includegraphics[scale=0.9]{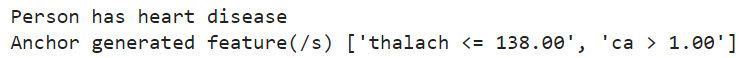}
\caption{Example : An output of Anchor}
\label{fig:anchors-example}
\end{figure}

\fakeparagraph{Example on Heart Disease Dataset} Let's take an example with Heart Disease dataset. Suppose we take an instance where the features of the person contribute to him/ her having a heart disease. The value of \texttt{thalach} of this person is $131$ and \texttt{ca} is $3$, therefore AnchorTabular method makes the following conclusions~(See Figure~\ref{fig:anchors-example}) .
We now apply the \texttt{AnchorTabular} method to see which features contribute significantly for such type of prediction.

\begin{verbatim}
target_label=['no heart disease','heart disease']
print('Person has',
target_label[explainer.predictor(x_test[instance].
             reshape(1, -1))[0]])
anchor=explainer.explain(x_test[instance])
print('Anchor generated feature(/s)',anchor.anchor)
\end{verbatim}

The person's maximum heart rate is $131$ (which is less than $138$). The blood vessels coloured by \texttt{fluoroscopy} are $3$ (greater than $1$).  As the maximum heart rate of a person should be high and the blood vessels coloured by \texttt{fluoroscopy} should be low, the above features act as anchors for the patient and deduce that the person has a heart disease.

\subsection{Counterfactuals}\label{sec:counterfactuals}
When a machine learning model is applied to real world data, along with the reason for the decision of its outcome, it is also essential to know  ``\textit{what should be the change in features in order to switch the prediction}''. Counterfactual explanations \cite{wachter2018counterfactual} is a model-agnostic XAI technique that provides the smallest changes that can be made to the feature values in order to change the output to the predefined output.

Counterfactual explainer methods work on black box models.\footnote{DiCE for Machine Learning:~\url{https://github.com/interpretml/DiCE}} Counterfactual explanations work best on binary datasets. They can also work on classification datasets with more than $3$ target values, but the performance is not as good as it is on binary classification datasets. In other words, if $X$ is an independent variable and $Y$ is a dependent variable, counterfactual shows the effect on $Y$ due to small changes in $X$. Also, it helps to calculate what changes need to be done in $X$ in order to change the outcome from $Y$ to $Y$'. It gives us the \textit{what-if} explanations for the model. Counterfactuals has a variety of applications in all sectors of society, from showing what changes need to be made to the feature values in order to change a rejected loan to an accepted loan to the changes that need to be made to the feature values of a patient suffering from a heart disease or breast cancer to change it from being fatal to non-fatal.

\fakeparagraph{An example on Heart Disease dataset}

\begin{verbatim}
m = dice_ml.Model(model=ann_model, backend=backend)
exp = dice_ml.Dice(d, m)
query_instance = {
    'age' : 67, 'sex' : 1, 'cp' : 3, 'trestbps' : 120, 
    'chol' : 229, 'fbs' : 0, 'restecg' : 2, 'thalach' : 129,
    'exang' : 1, 'oldpeak' : 2.6, 'slope' : 1 , 'ca' : 2,
    'thal' : 2 }
dice_exp = exp.generate_counterfactuals(query_instance
total_CFs=4
desired_class="opposite"
features_to_vary=['thalach', 'exang', 'oldpeak', 'slope',
'ca', 'thal', 'restecg', 'fbs','chol','trestbps'])
dice_exp.visualize_as_dataframe()
\end{verbatim}

\begin{figure*}[ht]
  \centering
  \includegraphics[scale=0.55]{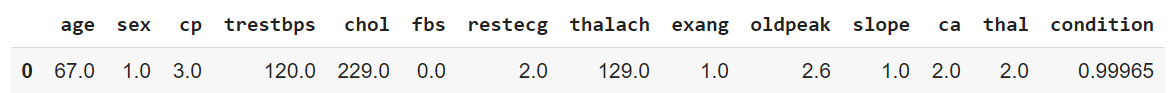}
  \caption{An input to a Counterfactual -- A specific instance where the patient has a heart disease.}
  \label{fig:cf-input}
\end{figure*}

The  value of \texttt{condition} field  $0$ signifies that the patient does not have a heart disease. The  value of \texttt{condition} field $1$  signifies that the patient does have a heart disease. From \textbf{Heart cancer dataset}, we have taken a specific instance where the patient has a heart disease. Figure~\ref{fig:cf-input} shows this instance.

\begin{figure*}[ht]
  \centering
  \includegraphics[scale=0.5]{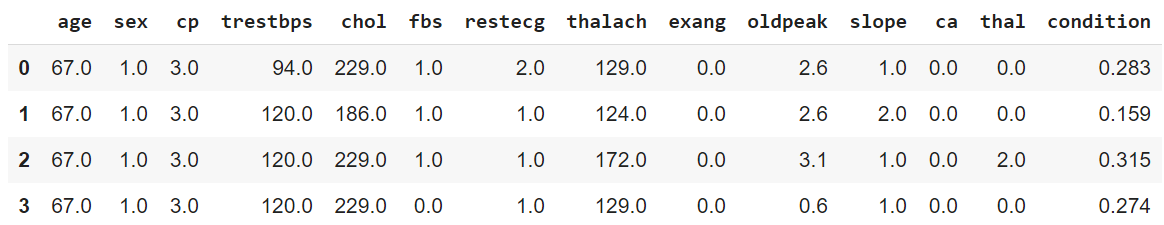}
  \caption{Different counterfactuals outputs of Counterfactual Explainer.}
  \label{fig:cf-output}
\end{figure*}

From the input presented in Figure~\ref{fig:cf-input}, We generate $4$ different counterfactuals as shown in Figure~\ref{fig:cf-output}, all of which show us the minimum changes that we can make to the feature values in order to change the condition of a patient. 

The following are the observations regarding the output:

\begin{itemize}
    \item We cannot change the \texttt{sex}, \texttt{age} or the type of chest-pain~(field \texttt{cp}) of a person suffering from a heart disease. Therefore, we can see that in each of the counterfactuals, we have kept those features unvaried. 

    \item The $4$ different counterfactuals are all different ways of solving one problem, which is to change the target value from $1$ to $0$. For example, if we look at the second counterfactual on the list, we can see that reduction of cholesterol will lead to decreasing the intensity of the heart disease. Among several other changes, it also shows that upon performing the Thallium test on the heart, there should be normal results and no defects.

    \item A recurring theme in all counterfactuals is the reduction of \texttt{ca} from \texttt{2} to \texttt{0}. \texttt{ca} signifies the number of blocked vessels of heart. \texttt{ca} is the most important feature contributing to having a heart disease. So, from the results, we can say that the most important factor in changing the condition is to reduce the number of blocked vessels by using methods like angioplasty.

\end{itemize}

\subsection{Counterfactuals with Prototypes}\label{sec:counterfactualswp}

It simply refers to the explanations that are described on the basis of a prototype i.e. a sample which is representative of the instances belonging to a class. Counterfactuals guided by prototypes is an advanced and more accurate version of counterfactuals \cite{vanlooveren2020interpretable}. The counterfactuals guided by prototypes method works on black-box models.This method is a model agnostic approach to interpret results using the prototypes of classes of target variable and is faster compared to the counterfactuals. It is much faster than counterfactual because of its prototype approach which speeds up the search process significantly by directing the counterfactual to the prototype of a particular class.

\begin{verbatim}
X = x_test[target].reshape((1,) + x_test[target].shape)
cf = CounterFactualProto(nn, shape, use_kdtree=True,
theta=10., max_iterations=1000,
feature_range=(x_train.min(axis=0), x_train.max(axis=0)), 
               c_init=1., c_steps=10)
cf.fit(x_train)
explanation = cf.explain(X)
print('Original prediction:
{}'.format(explanation.orig_class))
print('Counterfactual prediction:
{}'.format(explanation.cf['class']))
orig = X * sigma + mu
counterfactual = explanation.cf['X'] * sigma + mu
delta = counterfactual - orig
for i, f in enumerate(feature_names):
    if np.abs(delta[0][i]) > 1e-4:
      print('{}: {}'.format(f, delta[0][i]))
\end{verbatim}

\begin{figure*}[ht]
  \centering
  \includegraphics[scale=0.9]{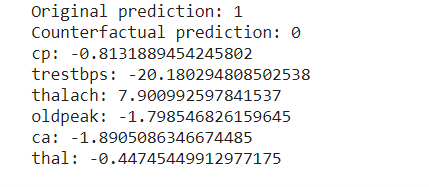}
  \caption{Example : An output of Counterfactual guided by prototype on heart disease dataset}
  \label{fig:cfprototype}
\end{figure*}

Here we take an instance where the condition of the patient is of having a heart disease. We now generate a counterfactual guided by prototype to see the smallest changes in the features this particular patient should make so that he can increase his/ her chances of being not diagnosed with a heart disease.

Figure~\ref{fig:cfprototype} shows that the following changes should be made to flip the outcome. The chest pain type should be decreased by |floor(-0.813)| = 1 i.e the chest pain should belong to the recognizable (symptomatic) category. Also, the resting blood pressure should be low. The maximum heart rate of the person should be higher by around 8, which can be done by regular exercise. Old peak value should be lowered by having a dash diet. ca should also be reduced by |floor(-1.89)| = 2 by detecting and removing the blockage using angioplasty or doing a bypass surgery if needed.

\subsection{Contrastive Explanation Methods}\label{sec:cem}

Contrastive Explanation Method, abbreviated as CEM, is a XAI Method which can give local explanations for a black box model. CEM defines explanations for classification models by providing insight on preferable along with unwanted features i.e. Pertinent Positives (PP) and Pertinent Negatives (PN). It is the first method that gives the explanations of both what should be minimally present and what should be necessarily absent from the instance to be explained in order to maintain the original prediction class.  In other words, the method finds the features like important pixels in an image that should be sufficiently present to predict the same class as on the original instance as well as identifies a minimal set of features that is enough to differentiate it from the nearest different class. 

The 2 kinds of explanations can be defined as follows~\cite{dhurandhar2018explanations}: 

\textbf{Pertinent Positives~(PP)}: The Pertinent Positives explanation finds the features that are necessary for the model to predict the same output class as the predicted class. For example, this includes the important pixels of an image, the features having a high feature weight, etc. PP works similarly to Anchors.

\textbf{Pertinent Negatives~(PN)}: The Pertinent Negatives explanation finds the features that should be minimally and sufficiently absent from an instance whilst maintaining the original output class. PN works similarly to Counterfactuals.

Using CEM, we can improve the accuracy of the machine learning model by looking at cases of mis-classified instances and then working on them using the explanations provided by CEM.

\begin{figure*}[ht]
\centering
\includegraphics[scale=0.47]{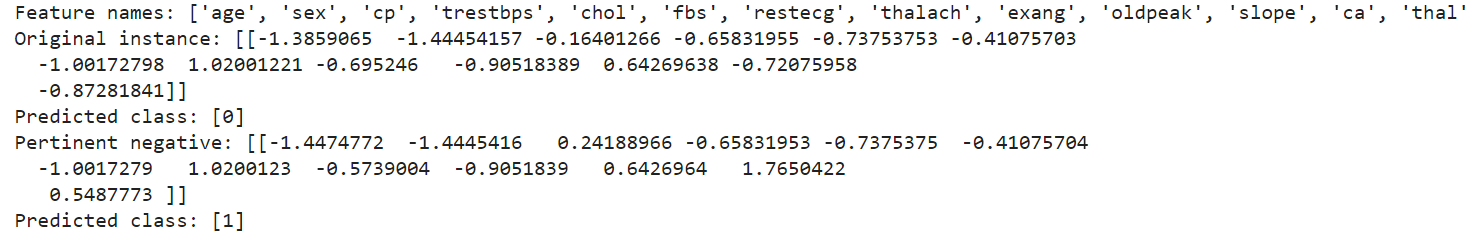}
\caption{Pertinent Negative Explanations generated by CEM on heart disease dataset.}
\label{fig:cem-output1}
\end{figure*}

\fakeparagraph{Pertinent Negative}
Figure~\ref{fig:cem-output1} generates contrastive explanations in terms of Pertinent Negative. The original prediction was $0$ which is changed to $1$ after applying CEM with pertinent negative. Pertinent Negative explanations work similarly to the counterfactual explanations, and we can see it clearly as the pertinent negative method pushes the prediction to get a prediction different from the original prediction which is $0$ to $1$ in this case.

\begin{verbatim}
idx = 1
X = x_test[idx].reshape((1,) + x_test[idx].shape)
mode = 'PN'
shape = (1,) + x_train.shape[1:]  
feature_range = (x_train.min(axis=0).reshape(shape)-.1,  
                 x_train.max(axis=0).reshape(shape)+.1)  
lr = load_model('nn_heart.h5')
cem = CEM(lr, mode, shape, kappa=kappa, beta=beta,
feature_range=feature_range, max_iterations=max_iterations,
c_init=c_init, c_steps=c_steps, learning_rate_init=lr_init,
clip=clip)
cem.fit(x_train, no_info_type='median')  
explanation = cem.explain(X, verbose=False)
print('Feature names: {}'.format(feature_names))
print('Original instance: {}'.format(explanation.X))
print('Predicted class: {}'.format([explanation.X_pred]))
print('Pertinent negative: {}'.format(explanation.PN))
print('Predicted class: {}'.format([explanation.PN_pred]))
\end{verbatim}

The CEM values in the array which are different from the original one change the prediction class. Some of them are \texttt{cp}, \texttt{ca}, \texttt{thal}. Thus, it can be concluded that changes in these features should necessarily be absent to retain the original prediction as $0$ as they are responsible for flipping the prediction class.

\begin{figure*}[ht]
  \centering
  \includegraphics[scale=0.62]{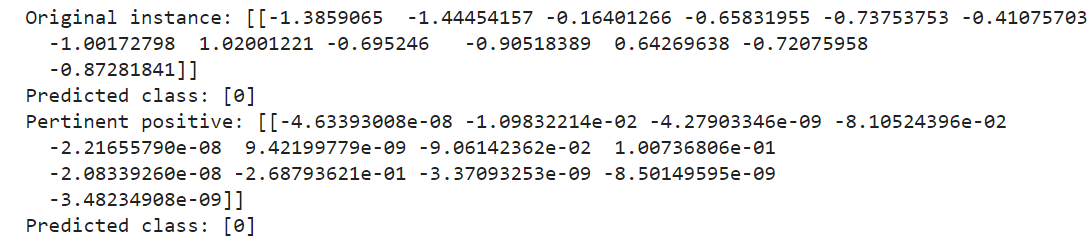}
  \caption{Pertinent Positive Explanations generated by CEM on heart disease dataset.}
  \label{fig:cem-output2}
\end{figure*}

\fakeparagraph{Pertinent Positive}
Generating CEM explanations in the form of PP shows us the feature values that should be compulsorily present in order to get the same original class ($0$) as predicted class, as shown in the example~(See Figure~\ref{fig:cem-output2}) . 

\begin{verbatim}
mode = 'PP'
lr = load_model('nn_heart.h5')
cem = CEM(lr, mode, shape, kappa=kappa, beta=beta,
feature_range=feature_range, max_iterations=max_iterations,
c_init=c_init, c_steps=c_steps, learning_rate_init=lr_init,
clip=clip)
cem.fit(x_train, no_info_type='median')
explanation = cem.explain(X, verbose=False)

print('Original instance: {}'.format(explanation.X))
print('Predicted class: {}'.format([explanation.X_pred]))
print('Pertinent positive: {}'.format(explanation.PP))
print('Predicted class: {}'.format([explanation.PP_pred]))
\end{verbatim}

The above result signifies that the predicted class remains the same on applying \texttt{PP}. The CEM values generated are close to $0$ as they are the most minimal values to be present for the prediction of the particular class i.e. these should be compulsorily and minimally present in order to get the same original class $0$ as predicted class.

\subsection{Kernel Shapley}\label{sec:kernelshap}

The goal of SHAP is to calculate the impact of every feature on the prediction~\cite{garcia2020shapley}. Compared to Shapley values, Kernel SHAP gives more computationally efficient and more accurate approximation in higher dimensions. In Kernel SHAP, instead of retraining models with different permutations of features to compute the importance of each, we can use the full model that is already trained, and replace "missing features" with "samples from the data" that are estimated from a formula. This means that we equate "absent feature value" with "feature value replaced by random feature value from data". Now, this changed feature space is fitted to the linear model and the coefficients of this model act as Shapley values.

It has the capability of both local and global interpretations i.e. it is able to compute the importance of each feature on the prediction for an individual instance and for the overall model as well \cite{lundberg2017unified}. The SHAP values are consistent and reliable because if a model changes so that the marginal contribution  of a feature value~(which means percentage out of the total) increases or stays the same (regardless of other features), they increase or remain the same respectively. Thus, these values are mathematically more accurate and require fewer evaluations. However, it assumes the features to be independent which may sometimes give wrong results.

\begin{verbatim}
pred = classifier.predict_proba
lr_explainer = KernelShap(pred, link='logit') 
lr_explainer.fit(X_train_norm)
\end{verbatim}

\begin{figure*}[ht]
  \centering
  \includegraphics[scale=0.6]{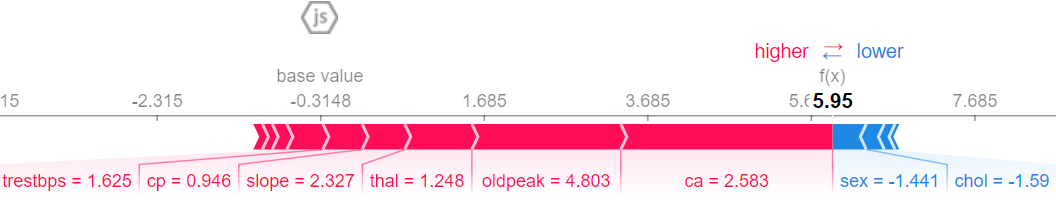}
  \caption{An example of local explanation using heart disease dataset.}
  \label{fig:kernel-shap-local}
\end{figure*}

\fakeparagraph{Local explanation} An example of local explanation using heart disease dataset is given in Figure~\ref{fig:kernel-shap-local}.
\begin{verbatim}
instance = X_test_norm[target][None, :]
pred = classifier.predict(instance)
class_idx = pred.item()
shap.force_plot(lr_explanation.expected_value[class_idx],
    lr_explanation.shap_values[class_idx][target,:], 
    X_test_norm[target][None, :],
    features_list)
\end{verbatim}

 The base value is the average of all output values of the model on the training data(here : -0.3148). Pink values drag/push the prediction towards $1$ (pushes the prediction higher i.e. towards having heart disease) and the blue towards $0$ (pushes the prediction lower i.e. towards no disease). The magnitude of influence is determined by the length of the features on the horizontal line. The value shown corresponding to the feature are the values of feature at the particular index(eg. $2.583$ for \texttt{ca}). Here, the highest influence is of \texttt{ca} for increasing the prediction value and of \texttt{sex} for decreasing the value.

\begin{figure*}[ht]
  \centering
  \includegraphics[scale=0.7]{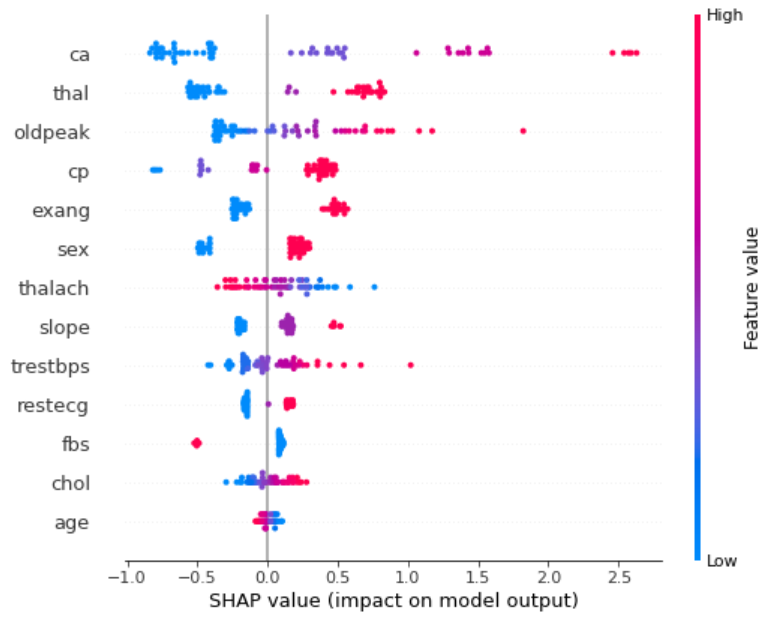}
  \caption{An example of global explanation using heart disease dataset.}
  \label{fig:kernel-shap-global}
\end{figure*}

\fakeparagraph{Global explanation} Figure~\ref{fig:kernel-shap-global} plots visualizes the impact of features on the prediction class $1$.
\begin{verbatim}
shap.summary_plot(lr_explanation.shap_values[1], 
                  X_test_norm, features_list)
\end{verbatim}

 The features are arranged such that the highest influence is of the topmost feature. Thus, \texttt{ca} is the feature that influences the prediction the most followed by \texttt{thal} and so on. The colour shades show the direction in which the feature impacts the prediction. For example, higher shap values of \texttt{ca} are shown in red colour which means high feature value. The higher the value of \texttt{ca}, the higher is the SHAP value i.e. more towards 1 . High value of \texttt{ca} indicates more chances of Heart Disease. However, it is the opposite for some features: High \texttt{thalach} will indicate no heart disease.

\subsection{Integrated Gradients}\label{sec:integratedgradients}

Integrated Gradients, also known as Path-Integrated Gradients or Axiomatic Attribution for Deep Networks is an XAI technique that gives an importance value to each feature of the input using the gradients of the model output~\cite{sundararajan2017axiomatic}. It is a local method that helps explain each individual prediction. Integrated Gradients is a method that is simple to implement and is applicable for various datasets such as text datasets, image datasets and structured data as well as models like regression and classification models. This method shows us the specific attributions of the feature inputs that are positive attributions and negative attributions.

\begin{itemize}
\item \textbf{Positive attributions: } Positive attributions are attributions that contributed or influenced the model to make the decision it did. For example, in the Fashion MNIST dataset, if we take the image of a shoe, then the positive attributions are the pixels of the image which make a positive influence to the model predicting that it is a shoe. It can be various specific pixels showing some unique features of a shoe such as laces, sole, etc. 
 
\item \textbf{Negative attributions: } Negative attributions are attributions that contributed or influenced the model against making the decision that it eventually did. For example, in the Fashion MNIST dataset, if we take the image of a shoe, then the negative attributions are the pixels of the image which go against the model predicting it as a shoe. For example, the sole of a shoe might be mixed as a flip flop. This would make the model think that the clothing item in the image is not a shoe. 

\end{itemize}

This method has various applications and is primarily used to check where the model is making mistakes so that amendments can be made in order to improve the accuracy of the model. In this way, it can help a developer in debugging as well as makes the model more transparent for the users. 

\fakeparagraph{An example on heart disease dataset} Let us understand it more using an example using MNIST dataset, as presented in  positive attributions and negative attributions.  We have taken an example on the dataset of Fashion MNIST, which consists of 70,000 grayscale 28x28 images of different clothing items. The label consists of $10$ classes, which denotes different kinds of clothing items such as shirt, hoodies, shoes and jeans.

\begin{verbatim}
n_steps = 50
method = "gausslegendre"
ig  = IntegratedGradients(model,
                          n_steps=n_steps,
                          method=method)
nb_samples = 10
X_test_sample = X_test[:nb_samples]
predictions = model(X_test_sample).numpy().argmax(axis=1)
explanation = ig.explain(X_test_sample,
                         baselines=None,
                         target=predictions)
explanation.meta
explanation.data.keys()
attrs = explanation.attributions
fig, ax = plt.subplots(nrows=2, ncols=4, figsize=(10, 7))
image_ids = [9, 4]
cmap_bound = np.abs(attrs[[9, 4]]).max()

for row, image_id in enumerate(image_ids):
    
    ax[row, 0].imshow(X_test[image_id].squeeze(), cmap='gray')
    ax[row, 0].set_title(f'Prediction:
    {predictions[image_id]}')
    attr = attrs[image_id]
    im = ax[row, 1].imshow(attr.squeeze(), 
    vmin=-cmap_bound, vmax=cmap_bound, cmap='PiYG')
    
    attr_pos = attr.clip(0, 1)
    im_pos = ax[row, 2].imshow(attr_pos.squeeze(),
    vmin=-cmap_bound, vmax=cmap_bound, cmap='PiYG')
    
    attr_neg = attr.clip(-1, 0)
    im_neg = ax[row, 3].imshow(attr_neg.squeeze(),
    vmin=-cmap_bound, vmax=cmap_bound, cmap='PiYG')

ax[0, 1].set_title('Attributions');
ax[0, 2].set_title('Positive attributions');
ax[0, 3].set_title('Negative attributions');

for ax in fig.axes:
    ax.axis('off')

fig.colorbar(im, cax=fig.add_axes([0.95, 0.25, 0.03, 0.5]));
\end{verbatim}

\begin{figure*}[ht]
  \centering
  \includegraphics[scale=0.6]{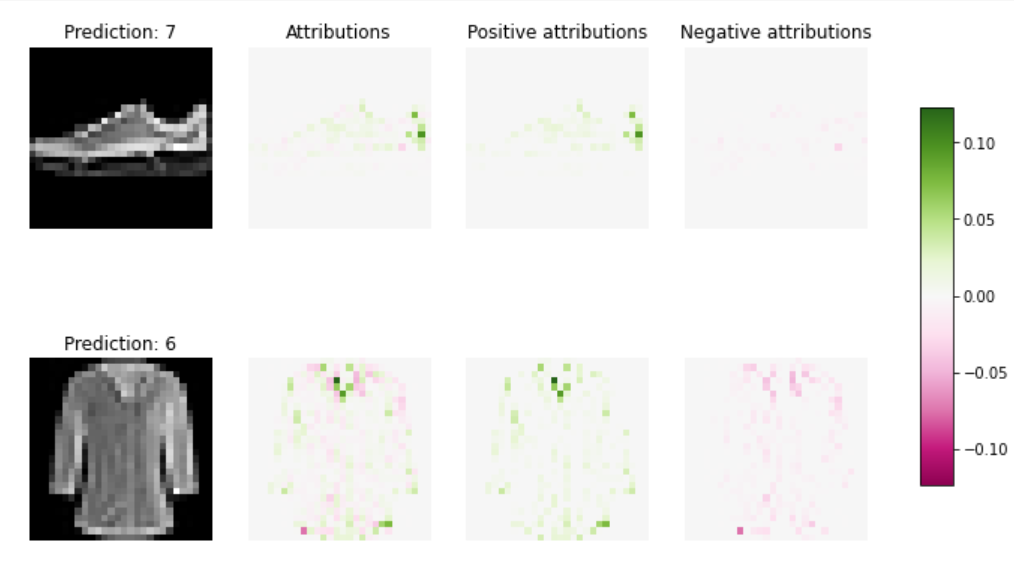}
  \caption{An example of Integrated Gradients using Fashion MNIST dataset.}
  \label{fig:ig}
\end{figure*}

The \textbf{first example}~(top part of Figure~\ref{fig:ig}) is is an image of a shoe. The attributions section shows a melange of positive and negative attributions together. As we can see from the bar on the right side, green signifies positive attributions and purple signifies negative attributions. The shoe is unique compared to other clothing items, and hence, it has a lot more positive attributes than negatives. The lining, collar and back part of the shoe are the main pixels that influence the decision of the model. On the other hand, the negative attributions are negligible for this particular instance. The \textbf{second example}~(bottom part of Figure~\ref{fig:ig}) is an image of a shirt where there is an equal number of positive and negative attributions. The pixels around the collar and the sleeves are the biggest positive attributions. However, the middle portion of the shirt can be mistaken to be a part of a pair of jeans or trousers. Therefore, due to this ambiguity, they are the negative attributions for the prediction. All in all, we can say that when the positive attributions outweigh the negative attributions, the model makes the correct prediction.

\section{Discussion}
In this paper, we have talked about different interpretability techniques. The aim is to enlighten practitioners on the understandability and interpretability of explainable AI systems using a variety of techniques available which can be very advantageous in the health-care domain. Medical diagnosis model is responsible for human life and we need to be confident enough to treat a patient as instructed by a black-box model. Our paper contains examples based on the heart disease dataset and elucidates on how the explainability techniques should be preferred to create trustworthiness while using deep learning systems in healthcare.

We have discussed the most popular feature based techniques LIME and SHAP which are very similar to each other in purpose but have very different approaches. Unlike LIME, which is only capable of local explanations, SHAP can explain both globally and locally. Multiple types of plots that explain the dataset globally are drawn using the dataset which provide information about relationships between the features and their importance. In a later section, a similar technique Kernel Shapley is discussed which is mathematically better and gives results in fewer evaluations. We then propose and discuss the different example based techniques namely Anchors, Counterfactuals, Integrated gradients, Contrastive Explanation Method, Kernel Shapley that play an integral role in revealing the black box nature for model transparency and accountability. Anchors, based on If-then rules, are better than LIME and SHAP in terms of generalisability and computation cost respectively. Counterfactual mainly presents the what-if explanations and its improved faster version Counterfactual with Prototype uses the prototypes of classes of target variable for these explanations. While, counterfactuals say how the features should change to change the target variable, CEM does the opposite and talks about what should necessarily be present to prevent the prediction to flip. CEM is a unique technique that provides Pertinent Positive - minimally present and Pertinent Negative - compulsorily absent features for original prediction class. The techniques so far only inform about how the features present give rise to the prediction and say nothing about the features that influenced to decide against the predicted class. This is what Integrated Gradients implement. Along with the positive attributions that help to make a prediction, it is the only feature which mentions the features that lead the model to believe a different prediction, known as negative attributions. Thus, the most important benefit learnt from the study of these techniques is that the approaches all speak how various features are responsible for the model’s outcomes. They are intuitive and thus assist in the process of learning what the black box model thinks and to be able to explain the behaviour of the model.

\bibliographystyle{splncs04}
\bibliography{arxivref}
\end{document}